\documentclass[conference]{IEEEtran}
\PassOptionsToPackage{numbers,square}{natbib}
\usepackage{natbib}

\IEEEoverridecommandlockouts
\usepackage{amsmath,amssymb,amsfonts}
\usepackage{algorithmic}
\usepackage{graphicx}
\usepackage{textcomp}
\usepackage{xcolor}
\usepackage{booktabs}
\usepackage{multirow}
\usepackage{colortbl}
\usepackage[numbers]{natbib}  
\usepackage[export]{adjustbox} 
\usepackage{float}

\def\BibTeX{{\rm B\kern-.05em{\sc i\kern-.025em b}\kern-.08em
    T\kern-.1667em\lower.7ex\hbox{E}\kern-.125emX}}
\begin{document}

\title{StripDet: Strip Attention-Based Lightweight 3D Object Detection from Point Cloud\\
\thanks{This work was supported in part by the Shenzhen Science and Technology
Program 2023A007; the Jiangsu Provincial Science and Technology Major Special Project BG2024032; and the Key Project of Shenzhen Basic Research Program JCYJ20241206180301003.\\
\hspace*{1em}Corresponding authors: Wendong Mao and Zhongfeng Wang.}
}
\author{\IEEEauthorblockN{Weichao Wang, Wendong Mao and Zhongfeng Wang}
\textit{School of Integrated Circuits, Shenzhen Campus of Sun Yat-sen University, China}\\
Email: wangwch23@mail2.sysu.edu.cn, maowd@mail.sysu.edu.cn, wangzf83@mail.sysu.edu.cn}

\maketitle

\begin{abstract}
The deployment of high-accuracy 3D object detection models from point cloud remains a significant challenge due to their substantial computational and memory requirements. To address this, we introduce StripDet, a novel lightweight framework designed for on-device efficiency. 
First, we propose the novel Strip Attention Block (SAB), a highly efficient module designed to capture long-range spatial dependencies. By decomposing standard 2D convolutions into asymmetric strip convolutions, SAB efficiently extracts directional features while reducing computational complexity from quadratic to linear.
Second, we design a hardware-friendly hierarchical backbone that integrates SAB with depthwise separable convolutions and a simple multi-scale fusion strategy, achieving end-to-end efficiency. Extensive experiments on the KITTI dataset validate StripDet's superiority. With only 0.65M parameters, our model achieves a 79.97\% mAP for car detection, surpassing the baseline PointPillars with a 7× parameter reduction. Furthermore, StripDet outperforms recent lightweight and knowledge distillation-based methods, achieving a superior accuracy-efficiency trade-off while establishing itself as a practical solution for real-world 3D detection on edge devices.
\end{abstract}


\section{Introduction} 
With the rapid advancement of autonomous driving and intelligent robotics, robust real-time 3D scene understanding has become increasingly critical. Among various perception tasks, 3D object detection from LiDAR point clouds serves as a fundamental component for environmental awareness, enabling real-world applications including autonomous vehicles \cite{song2024robustness}, virtual reality systems \cite{ghasemi2022deep} and mobile robotics platforms \cite{zhang2024towards}.
Building on deep learning techniques and large-scale datasets \cite{geiger2013vision}, substantial progress has been achieved in developing specialized architectures for point cloud processing \cite{li2023pillarnext}. However, these advancements are often accompanied by increased computational complexity and considerable parameter overhead, posing significant challenges for deployment in resource-constrained environments.

The need to mitigate high computational complexity and substantial parameter overhead has spurred significant research into lightweight 3D object detection frameworks. These frameworks seek to establish an optimal trade-off between detection accuracy and computational efficiency.
To this end, a spectrum of model compression strategies has been explored, including network pruning, quantization, knowledge distillation \cite{zhang2023pointdistiller}, and lightweight architecture design \cite{howard2019searching}.      
However, prevailing lightweight approaches are hampered by a fundamental trade-off. For instance, methods like PointDistiller \cite{zhang2023pointdistiller} rely on computationally expensive operations and complex teacher-student paradigms, which limits their practical utility. Conversely, other approaches \cite{li2023lightweight} that simplify core 3D operations often suffer performance degradation, necessitating a dependency on knowledge distillation that introduces training complexity. 
This observation underscores a critical research gap: the need for architectures that are intrinsically efficient without compromising representational capacity.
To address this gap, we introduce StripDet, a lightweight and effective 3D object detection framework. We fundamentally reconstruct the backbone of the conventional pillar-based architecture by introducing the Strip Attention Block (SAB). This novel component utilizes asymmetric strip-shaped convolutions to capture long-range, anisotropic spatial dependencies with minimal computational overhead. This architectural innovation enables highly discriminative feature modeling while drastically reducing model size. The key contributions of this work are summarized as follows:
\begin{itemize}
\item We propose the Strip Attention Block, a novel and efficient unit built upon asymmetric strip convolutions. The SAB is designed to capture long-range, anisotropic spatial dependencies in point cloud features while maintaining minimal computational overhead.
\item We introduce StripDet, a lightweight, end-to-end 3D object detection framework. By fundamentally reconstructing the backbone of a conventional pillar-based model with our proposed SABs, StripDet delivers high detection accuracy while maintaining exceptional model efficiency.
\item Comprehensive experiments on the KITTI dataset demonstrate that StripDet outperforms recent lightweight and knowledge distillation-based approaches. Specifically, our model achieves a highly competitive \textbf{79.97\%} mAP for car detection with only \textbf{0.65M} parameters, surpassing the PointPillars~\cite{lang2019pointpillars} baseline with a \textbf{7×} parameter reduction.
\end{itemize}

\vspace{0.2cm} 
\section{Proposed Method}
This section details the proposed lightweight 3D object detection framework, StripDet. We begin by introducing the Strip Attention Module (SAM) and the Strip Attention Block (SAB), which form the fundamental building blocks of our network. We then present the overall architecture, detailing how these components are integrated and outlining the design principles that guide its construction.
\subsection{Preliminaries}
We consider an input LiDAR point cloud as a set of points $ \mathcal{X} = \{\mathbf{x}_1, \mathbf{x}_2, \dots, \mathbf{x}_n\} $, where each point $ \mathbf{x}_i \in \mathbb{R}^4 $ comprises 3D coordinates $ (x, y, z) $ and an intensity value. The objective of 3D object detection is to predict a set of 3D bounding boxes $ \mathcal{Y} $, each with an associated class label and confidence score. This task can be formulated as learning a function $ F = g \circ f $, where a feature extractor $ f: \mathcal{X} \rightarrow \mathcal{F} $ maps the point cloud to a high-dimensional representation $ \mathcal{F} $, and a detection head $ g: \mathcal{F} \rightarrow \mathcal{Y} $ generates the final predictions. In grid-based methods, a paradigm adopted by StripDet, the feature extractor $f$ first transforms the unstructured point cloud into a structured bird's-eye-view (BEV) feature map, denoted as $ \mathbf{f}(x) \in \mathbb{R}^{H \times W \times C} $, where $ H $ and $ W $ are the height and width of the BEV grid, and $ C $ is the number of feature channels.

\subsection{Strip Attention Module and Block Design}

The sparse and irregular nature of 3D point clouds poses a fundamental challenge for object detection. This sparsity causes objects to manifest as disconnected point clusters, making it crucial for models to capture long-range contextual information to recognize them as coherent entities. Consequently, a large receptive field is not merely beneficial but essential for robust feature learning, a point underscored by recent work \cite{li2023pillarnext}. 
However, conventional approaches to expand the receptive field, such as stacking small convolutions \cite{yan2018second} or employing large-kernel convolutions \cite{zhang2024scaling} and Transformer blocks \cite{kolodiazhnyi2024oneformer3d}, are often computationally prohibitive.This inherent trade-off between receptive field size and efficiency severely limits their applicability in real-time scenarios, motivating the need for a more resource-conscious architectural design.

\begin{figure}[htbp]
\vspace{-2pt}
\centering
\includegraphics[scale=0.8]{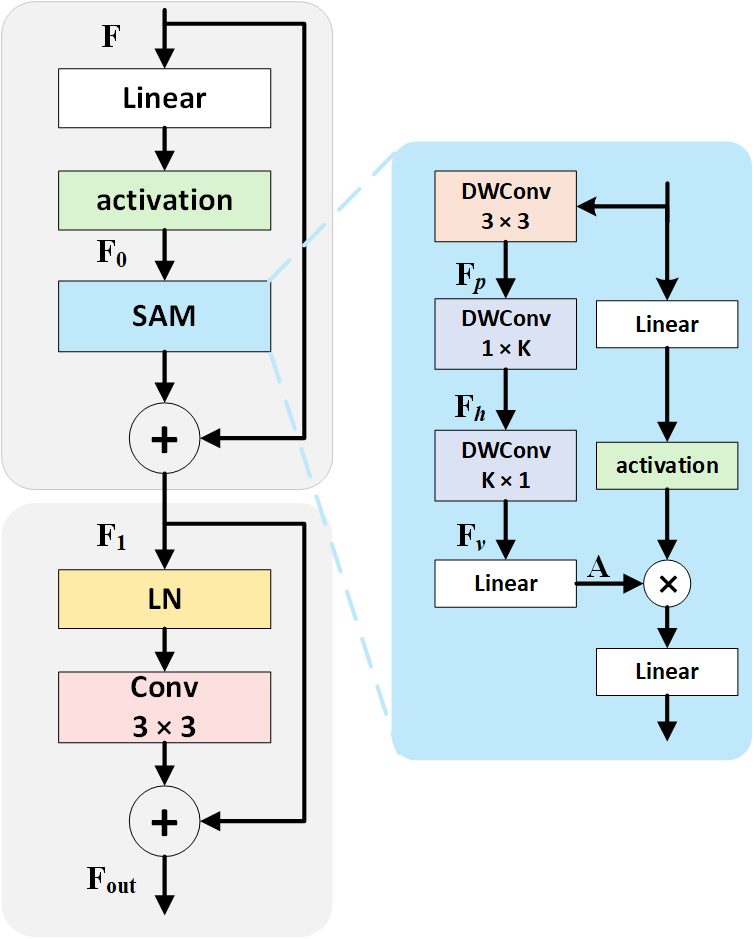}
\vspace{-2pt}
\caption{Architecture of the Strip Attention Module (SAM) and Strip Attention Block (SAB).}
\label{SAB}
\vspace{-2pt}
\end{figure}

To resolve the tension between representational capacity and efficiency, we first introduce the Strip Attention Module (SAM). As illustrated in Fig. \ref{SAB}, SAM is engineered to enhance directional feature modeling and efficiently expand the receptive field. Given an input feature map $ \mathbf{F}_{0} \in \mathbb{R}^{H \times W \times C} $, the module begins by applying a 3×3 depthwise convolution to project and refine the features:
\begin{equation}
    \mathbf{F}_{p} = \text{DWConv}_{3 \times 3}(\mathbf{F}_{0}).
\end{equation}
Subsequently, two depthwise convolutions are applied sequentially: one along the horizontal direction ($1 \times K$) and the other along the vertical direction ($K \times 1$) to capture long-range contextual dependencies along specific spatial dimensions with minimal computational overhead:
\begin{align}
    \mathbf{F}_{h} &= \text{DWConv}_{1 \times K}(\mathbf{F}_{p}), \\
    \mathbf{F}_{v} &= \text{DWConv}_{K \times 1}(\mathbf{F}_{h}).
\end{align}
This sequential design enables progressive aggregation of directional context, forming a lightweight yet effective attention mechanism.

Next, the refined directional feature $\mathbf{F}_{v}$ is fed into a pointwise convolution to generate a spatial attention map:
\begin{equation}
    \mathbf{A} = \text{PWConv}(\mathbf{F}_{v}).
\end{equation}
Meanwhile, the original input feature $\mathbf{F}_{0}$ is refined via a linear projection and then activated through a GeLU nonlinearity. The final feature map is obtained by performing element-wise multiplication between the activated input and the spatial attention map:

 \begin{equation} 
    \mathbf{F}' = \text{GeLU}(\text{Linear}(\mathbf{F}_{0})) \odot \mathbf{A}. 
\end{equation}
Building upon this module, we construct the SAB, which serves as the fundamental building block of our backbone. As illustrated in Fig. \ref{SAB}, the SAB consists of two residual sub-blocks: an SAM-guided sub-block and a feed-forward network sub-block.
Formally, given an input feature map $\mathbf{F} \in \mathbb{R}^{H \times W \times C}$, the first component applies channel-wise transformation followed by nonlinearity and spatial attention:
\begin{equation}
    \mathbf{F}_1 = \mathbf{F} + \text{SAM}\left( \text{GeLU} \left( \text{Linear}(\mathbf{F}) \right) \right),
\end{equation}
where $\text{Linear}(\cdot)$ denotes a linear layer, used for channel projection, and $\text{SAM}(\cdot)$ refers to the Strip Attention Module.
The second component further enhances local contextual modeling through a normalization layer and a standard convolution:
\begin{equation}
    \mathbf{F}_{\text{out}} = \mathbf{F}_1 + \text{Conv}_{3\times3} \left( \text{LayerNorm}(\mathbf{F}_1) \right).
\end{equation}
Here, LayerNorm stabilizes internal representations across channels, while the $3\times3$ convolution captures fine-grained spatial dependencies. Both operations are performed within a residual framework to preserve information from earlier layers.

Notably, the SAB's minimalist design dispenses with complex attention mechanisms by decoupling 2D spatial attention into orthogonal 1D strips, thereby achieving a superior balance between representational capacity and computational cost.
\subsection{Overall Architecture}
\begin{figure*}[t]
    \centering
    \includegraphics[width=\textwidth, max height=0.2\textheight, keepaspectratio, valign=c]{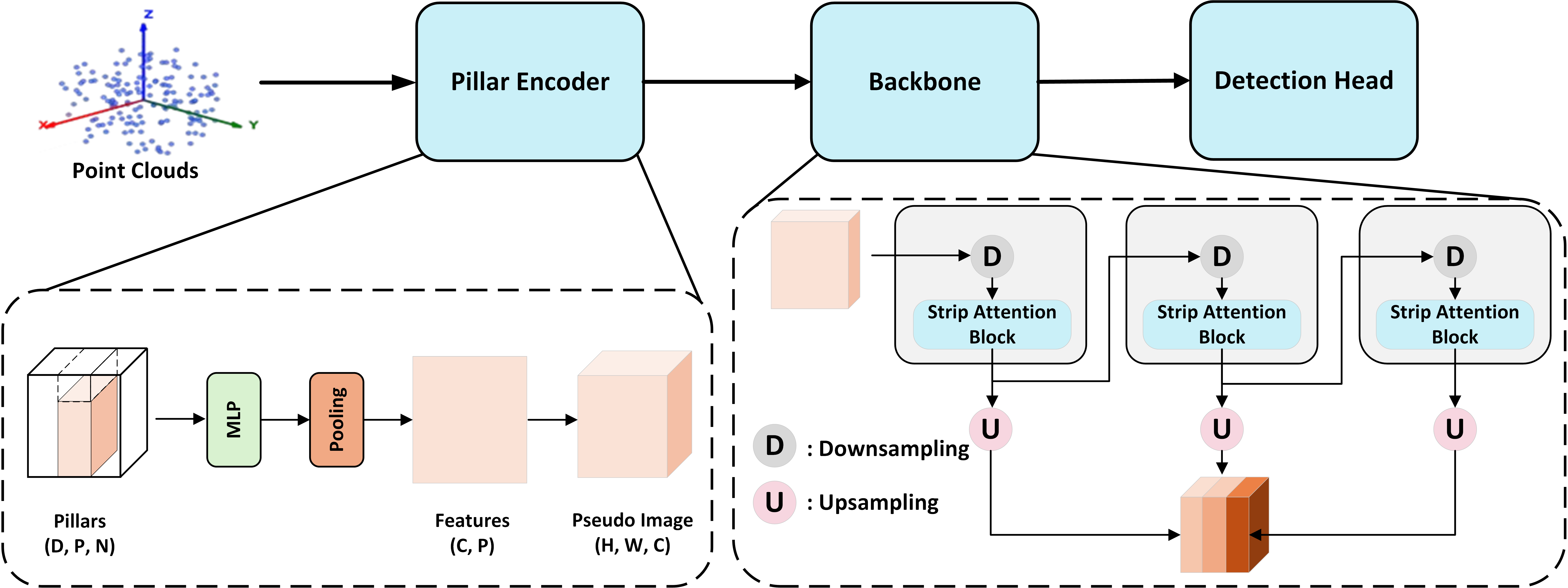}
    \caption{An overview of the proposed StripDet network architecture.}
    \label{network}
    \vspace{-6pt}
\end{figure*}

As illustrated in Fig.~\ref{network}, the pipeline consists of three core components: a \textit{Pillar Encoder}, a \textit{Hierarchical Backbone} constructed from our proposed SAB, and a \textit{Detection Head}.

\textbf{Pillar Encoder:} Following the pillar-based paradigm, the encoder transforms unstructured LiDAR point clouds into a structured BEV feature map by grouping points into vertical pillars. Each pillar aggregates local geometric and intensity information, producing an initial feature representation with dimensions $H \times W \times C_0$, where $C_0=64$.

\textbf{Hierarchical Backbone:} The backbone's remarkable efficiency is achieved through a synergistic combination of our novel SAB and proven efficient techniques. As our core innovation, the SAB is engineered to efficiently replace computationally expensive mechanisms such as standard 2D convolutions and attention. Its efficiency stems from a spatial factorization that decomposes a 2D operation into two sequential, asymmetric 2D convolutions (with $1 \times K$ and $K \times 1$ kernels). This approach provides a twofold advantage. First, it reduces the computational complexity with respect to the kernel size from quadratic ($O(K^2)$) to linear ($O(K)$), leading to a drastic reduction in parameters and FLOPs. Second, and arguably more critical for practical inference speed, this factorization enables highly efficient memory access. Specifically, the horizontal ($1 \times K$) convolution operates on contiguous data, which allows for coalesced memory reads that drastically reduce latency. To further enhance efficiency, we employ depthwise separable convolutions for spatial downsampling between stages. Finally, the multi-scale features from all three stages are fused via simple upsampling and concatenation, deliberately avoiding complex and parameter-heavy feature pyramid networks.

\textbf{Detection Head:} The detection head consists of three parallel branches for classification, bounding box regression, and orientation prediction. Each branch employs standard convolutions for task-specific output generation.

\vspace{-0.2cm} 
\subsection{Training Details}
We train the network end-to-end using a multi-task loss function that jointly optimizes object classification, localization, and orientation. The total loss $\mathcal{L}$ is a weighted sum of three components:
\begin{equation}
    \mathcal{L} = \mathcal{L}_{\text{cls}} + 2.0 \cdot \mathcal{L}_{\text{bbox}} + 0.2 \cdot \mathcal{L}_{\text{dir}},
\end{equation}
where $\mathcal{L}_{\text{cls}}$, $\mathcal{L}_{\text{bbox}}$, and $\mathcal{L}_{\text{dir}}$ are the Focal Loss, Smooth L1 loss, and cross-entropy loss, used for addressing class imbalance, robust bounding box regression, and orientation classification, respectively. The model is trained using the AdamW optimizer with a one-cycle learning rate policy.

\vspace{0.1cm} 
\section{Experiments and Results}
\subsection{Experimental Setup}

We evaluate our method on the KITTI dataset \cite{geiger2013vision}, using the standard split of 3,712 training and 3,769 validation samples. For evaluation, we report the mean average precision (mAP) with 40 recall positions on the validation set for both BEV and 3D detection. Following the official KITTI evaluation protocol, the Intersection-over-Union (IoU) threshold is set to 0.7 for the car class and 0.5 for pedestrian and cyclist.

\subsection{Comparison with Lightweight and General Purpose Detectors}
We compare StripDet with both lightweight and general-purpose 3D object detection methods in terms of detection accuracy, model size (parameters), and computational cost (FLOPs).

\begin{table}[h]
\vspace{-8pt}
\renewcommand{\arraystretch}{1.1}
    \centering
    \caption{Performance comparison with lightweight detectors on the KITTI validation set for BEV(Bird-Eye-View) and 3D tasks.}  
    \label{tab:lightweight_comparison}
    \setlength{\tabcolsep}{4pt} 
    \begin{tabular}{l l c c c c c}
        \hline
        \multirow{2}{*}{Task} & \multirow{2}{*}{Model} & \multicolumn{1}{c}{Car} & \multicolumn{1}{c}{Pedestrian} & \multicolumn{1}{c}{Cyclist} & \multirow{2}{*}{F (G)} & \multirow{2}{*}{P (M)} \\
        \cline{3-5}
        & & Mod. & Mod. & Mod. & \\
        \hline
        \multirow{5}{*}{BEV} & SECOND \cite{yan2018second} & 79.4 & 46.3 & 56.0 & 69.8 & 5.34 \\
        & PointPillars \cite{lang2019pointpillars} & 88.1& 51.8	& 65.0 &  34.3 & 4.83\\
        & LW-S-3 \cite{li2023lightweight} & 79.1 & 45.7 & 56.1 & 18.5 & 0.67\\
        & PointDistiller \cite{zhang2023pointdistiller} & \textbf{89.0} & 52.8 & 65.8 & \textbf{9.0} & 1.30\\
        \rowcolor{gray!20} & StripDet(ours) & 88.2 & \textbf{56.2} & \textbf{69.4 }& 9.5 & \textbf{0.65}\\
        \hline
        \multirow{5}{*}{3D} & SECOND \cite{yan2018second} & 73.7 & 42.6 & 53.9 & 69.8 & 5.34 \\
        & PointPillars \cite{lang2019pointpillars} & 75.9 & 45.9 & 59.2  &  34.3 & 4.83\\
        & LW-S-3 \cite{li2023lightweight} & 73.4 & 41.9 & 53.6 & 18.5 & 0.67 \\
        & PointDistiller \cite{zhang2023pointdistiller} & 76.9 & 47.5 & 62.0 & \textbf{9.0} & 1.30\\
        \rowcolor{gray!20} & StripDet(ours) & \textbf{78.1} & \textbf{51.1} & \textbf{65.3} & 9.5 & \textbf{0.65}\\
        \hline
        \multicolumn{7}{l}{\parbox[t]{\dimexpr\linewidth-2\tabcolsep\relax}{\footnotesize Note: All results are Average Precision (AP) for the Moderate difficulty, evaluated at 40 recall positions. \textbf{F} and \textbf{P} indicate FLOPs and parameters, respectively.}}
    \end{tabular}
\vspace{-6pt}
\end{table}

\begin{table*}[!t]
\centering
\renewcommand{\arraystretch}{1.3}
\caption{Performance comparison with general-purpose detectors on the KITTI validation set for 3D object detection. Best in \textbf{bold}.}
\label{tab:comparison}
\resizebox{\textwidth}{!}{%
\begin{tabular}{l|c|cccc|cccc|cccc|c}
\toprule
\multirow{2}{*}{\raisebox{-0.5\height}{Method}} & \multirow{2}{*}{\raisebox{-0.5\height}{Modality}} & \multicolumn{4}{c|}{Car} & \multicolumn{4}{c|}{Pedestrian } & \multicolumn{4}{c|}{Cyclist} & \multirow{2}{*}{\raisebox{-0.5\height}{Params (M)}} \\
\cmidrule(lr){3-6} \cmidrule(lr){7-10} \cmidrule(lr){11-14}
       &          & Easy & Mod. & Hard & Avg & Easy & Mod. & Hard & Avg & Easy & Mod. & Hard & Avg & \\
\midrule
VoxelNet \cite{zhou2018voxelnet} & L & 82.0 & 65.5 & 62.8 & 70.13 &  57.9 & 53.4 & 48.9 & 53.40 & 67.2 & 47.7 & 45.1 & 53.33 & 8.35 \\
PointPillars \cite{lang2019pointpillars} & L & 87.3 & 75.9 &  71.1 & 78.10 &  52.0 &  45.9 &  41.4 & 46.43 &  78.6 & 59.2 & 55.8 & 64.53 &  4.83 \\
PointRCNN \cite{shi2019pointrcnn} & L & 86.8 & 76.1 & \textbf{74.3} & 79.06  & 63.3 & 58.3 & 51.6 & 57.73 & \textbf{83.7} & \textbf{66.7} & \textbf{61.9} & \textbf{70.77} & 4.04 \\
SeSame + pillar\cite{hayeon2024sesame} & L &  85.7 & 75.6 & 73.9 & 78.40 & 53.4 & 48.4 & 44.7 & 48.83 &  77.9 & 61.4 & 58.3 & 65.87 & - \\
\midrule
MV3D \cite{chen2017multi} & L+I & 71.3 & 62.7 & 56.6 & 63.53 & - & - & - & - & - & - & - & - & - \\
AVOD-FPN \cite{ku2018joint} & L+I & 84.4 & 74.4 & 68.7 & 75.83 & - & 58.8 & - & - & - & 49.7 & - & - & 38.07 \\
PointFusion \cite{xu2022int} & L+I & 77.9 & 63.0 & 53.3 & 64.73 & 33.4 & 28.0 & 23.4 & 28.27 & 49.3 & 29.4 & 26.9 & 35.20 & - \\
F-PointNet \cite{qi2018frustum} & L+I & 83.8 & 70.9 & 63.6 & 72.76 & \textbf{70.0} & \textbf{61.3} & \textbf{53.6} & \textbf{61.63} & 77.1 & 56.4 & 53.3 & 62.27 & 22.00 \\
\midrule
PointDistiller \cite{zhang2023pointdistiller} & L & 88.1 & 76.9 & 73.8 & 79.60 & 54.6 & 47.5 & 42.3 & 48.13 & 80.3 & 62.0 & 58.8 & 67.03 & 1.30 \\
LW-S-3 \cite{li2023lightweight} & L & 82.8 & 73.4 & 65.2 & 73.80 &	51.4 & 41.9 & 37.7 & 43.67 &	69.5 &	53.6 &	45.8 & 56.30 & 0.67 \\
\rowcolor{gray!20}StripDet (Ours) & L & \textbf{88.1} & \textbf{78.1} & 73.7 & \textbf{79.97} & 57.8 & 51.1 & 46.5 & 51.80 & 83.4 & 65.3 & 61.6 & 70.10 & \textbf{0.65} \\
\bottomrule
\multicolumn{15}{l}{\parbox{\linewidth}{\footnotesize Note: All results are Average Precision (AP, \%) at 40 recall positions. Mod. = Moderate difficulty; Avg = Mean of Easy, Mod., and Hard APs. Modality: L = LiDAR, L+I = LiDAR+Image.}} \\
\end{tabular}%
}
\vspace{-2pt}
\end{table*}

As shown in Table~\ref{tab:lightweight_comparison}, StripDet significantly outperforms the baseline PointPillars~\cite{lang2019pointpillars} in both efficiency and accuracy. It reduces the model parameters by 7× and FLOPs by 3.6×, while simultaneously improving performance on challenging classes, achieving a 4.4\% mAP gain for pedestrian (BEV) and 6.2\% for cyclist (3D). When compared to the lightweight method LW-S-3~\cite{li2023lightweight}, StripDet demonstrates a superior accuracy-efficiency trade-off. With less than half the computational cost (FLOPs), StripDet surpasses LW-S-3 by a large margin, achieving +9.1\% and +4.7\% mAP gains in BEV and 3D car detection, respectively. Furthermore, we compare StripDet with the knowledge distillation-based method PointDistiller~\cite{zhang2023pointdistiller}. Despite using only half the parameters and a much simpler training paradigm without a teacher model, StripDet consistently outperforms PointDistiller, achieving significant mAP gains across all categories in 3D detection (+1.2\% for car, +3.6\% for pedestrian, and +3.3\% for cyclist). This highlights that StripDet's intrinsic architectural efficiency is a more effective approach than relying on complex, resource-intensive knowledge distillation techniques.

In addition, we compare StripDet with general-purpose detectors on the KITTI validation set, as shown in Table~\ref{tab:comparison}. The results highlight StripDet's exceptional ability to deliver high accuracy with extreme efficiency. With only 0.65M parameters, StripDet achieves highly competitive performance across all categories.
Remarkably, for car detection, StripDet achieves the highest mAP of 79.97\%, surpassing all listed general-purpose detectors. For pedestrian and cyclist detection, StripDet further demonstrates its effectiveness, leading all lightweight designs and remaining highly competitive with heavyweight detectors.
This consistent and balanced performance across all object classes and difficulty levels, achieved with a fraction of the parameters and computational cost, directly underscores the superiority of our hardware-friendly architecture. 

Collectively, the experimental results demonstrate that StripDet delivers both high accuracy and exceptional efficiency for lightweight 3D object detection. Its consistent outperformance of both lightweight and general-purpose detectors underscores the effectiveness of our architectural innovations. These findings position StripDet as a practical and scalable solution for real-world 3D perception tasks, especially in resource-constrained scenarios.

\vspace{0.3cm} 
\section{Conclusion}

In this paper, we introduced StripDet, a lightweight 3D object detection framework designed to resolve the persistent trade-off between accuracy and on-device efficiency. A key component of our approach is the Strip Attention Block, which uses asymmetric strip convolutions to efficiently capture long-range spatial dependencies with minimal computational overhead.
Extensive experiments on the KITTI dataset demonstrate the superiority of our design. Notably, StripDet achieves a 79.97\% mAP for car detection with only 0.65M parameters, significantly outperforming existing lightweight methods while remaining competitive with much larger, resource-intensive models. These results highlight StripDet's suitability for real-world applications on resource-constrained platforms.

\bibliographystyle{IEEEtran}  
\bibliography{ref}     %

\end{document}